\documentclass{article}

\usepackage[preprint]{nips_2018}

\usepackage[utf8]{inputenc} 
\usepackage[T1]{fontenc}    
\usepackage{hyperref}       
\usepackage{url}            
\usepackage{booktabs}       
\usepackage{amsfonts}       
\usepackage{nicefrac}       
\usepackage{microtype}      
\usepackage{amsmath}
\usepackage{amssymb}
\usepackage{graphicx}
\usepackage[table]{xcolor}
\usepackage{subfigure}
\usepackage{multirow}
\title{Spherical Convolutional Neural Network \\ for  3D Point Clouds}

\author{
  Huan Lei, Naveed Akhtar and Ajmal Mian\\
  Computer Science and Software Engineering\\
  University of Western Australia\\
  35 Stirling Highway, Crawley 6009, WA. Australia. \\
  \texttt{huan.lei@research.uwa.edu.au, \{naveed.akhtar, ajmal.mian\}@uwa.edu.au} \\
}

\begin{document}

\maketitle
\begin{abstract}
We propose a neural network for  3D point cloud processing that exploits \emph{spherical} convolution kernels and octree partitioning of space. The proposed metric-based spherical kernels systematically quantize point neighborhoods to identify local geometric structures in data, while maintaining the properties of translation-invariance and asymmetry. The network architecture itself is guided by octree data structuring that  takes full advantage of the sparse nature of irregular point clouds. We specify spherical kernels with the help of neurons in each layer that in turn are associated with spatial locations. We exploit this association to avert dynamic kernel generation during network training, that enables  efficient learning with high resolution point clouds. 
We demonstrate the utility of the spherical convolutional neural network for 3D object classification on standard benchmark datasets.


\end{abstract}
\vspace{-5mm}
\section{Introduction}
\label{sec:Intro}
\vspace{-2mm}
Convolutional Neural Networks (CNNs) [6] are known to learn  highly effective features from data. However, the standard CNNs are only amenable to data defined over regular grids, e.g. pixel arrays. This  limits their ability in processing 3D point clouds that are inherently irregular. Point clouds have recently gained significant interest of computer vision community and large repositories for this data modality have started to emerge~[15], [19].
Correspondingly, the related literature has also seen attempts to exploit the representation prowess of standard convolutional networks for point clouds by adaption [2], [8], [15]. However, these  attempts also result in excessively large memory footprints that restrict the allowed input data resolution [12].
A more attractive choice is to combine the power of convolution operation  with graph representations of irregular data.
The resulting Graph Convolutional Networks (GCNs) can offer convolutions either in the spectral domain [1], [4], [20] or the spatial domain [12].

In GCNs, the spectral domain methods require the Graph Laplacian to be aligned, which is not straight forward to achieve for point clouds. On the other hand, the only prominent approach in the spatial domain is ECC [12] that, in contrast to the standard CNNs, must generate convolution kernels dynamically which  entails a significant computational overhead. Additionally, ECC relies on range searches for graph construction and coarsening, that can become prohibitively expensive for large point clouds. 
One major challenge in applying convolutional networks to irregular 3D data is in specifying geometrically meaningful convolution kernels in the 3D metric space.
Naturally, the kernels are also required to exhibit translation-invariance to identify similar local structures in data. Moreover, they should be applied to point pairs asymmetrically for a compact representation.   
Owing to such intricate requirements, there are also few instances that altogether avoid the use of convolution kernels in computational graphs to process  unstructured data [5], [9], [10]. Although still attractive, these methods do not contribute towards harnessing the potential power of convolutional networks for point clouds. 

In this work, we introduce the notion of spherical convolutional kernel that systematically partitions a spherical 3D region into multiple volumetric bins, see Fig.~\ref{fig:sphConv}. Each bin specifies a matrix of learnable parameters that weights the points falling within that bin for convolution.
We apply these kernels between the layers of a neural network that is constructed based on octree partitioning [21] of the 3D space. The sparsity guided octree structuring determines the locations to perform the convolutions in each layer of our network. The network architecture itself is guided by octree hierarchy, having the same number of hidden layers as the tree depth. By exploiting space partitioning, the network avoids K-NN/range search and efficiently consumes high resolution point clouds. The spherical  kernels, that do not require dynamic generation, are also able to share weights between similar local structures in data. Moreover, we demonstrate that the kernels apply asymmetrically to point pairs in our network.  
We evaluate the spherical convolutional neural network for the task of 3D object recognition, achieving highly encouraging results for a novel concept of convolutional networks.

\vspace{-2mm}
\section{Spherical Convolutions} \label{sec:SC}
\vspace{-2mm}

For images, hand-crafted features have traditionally been computed over more primitive constituents, i.e.~patches. In effect, the same principle transcended to the automatic feature extraction with CNNs that compute feature maps using the activations of well-defined rectangular regions of images. Whereas rectangular regions are the common choice to process the data of 2D nature, spherical regions are more suited for processing 3D data such as point clouds. Spherical regions are inherently amenable to computing geometrically meaningful features for unstructured data [3], [13]. 
Inspired by this natural kinship, we introduce the concept of \emph{spherical convolution} that considers a 3D sphere (instead of a 2D rectangle) as the basic geometric shape to perform the convolution operation. 

 

\begin{figure}[t]
  \centering
  \includegraphics[height=35mm]{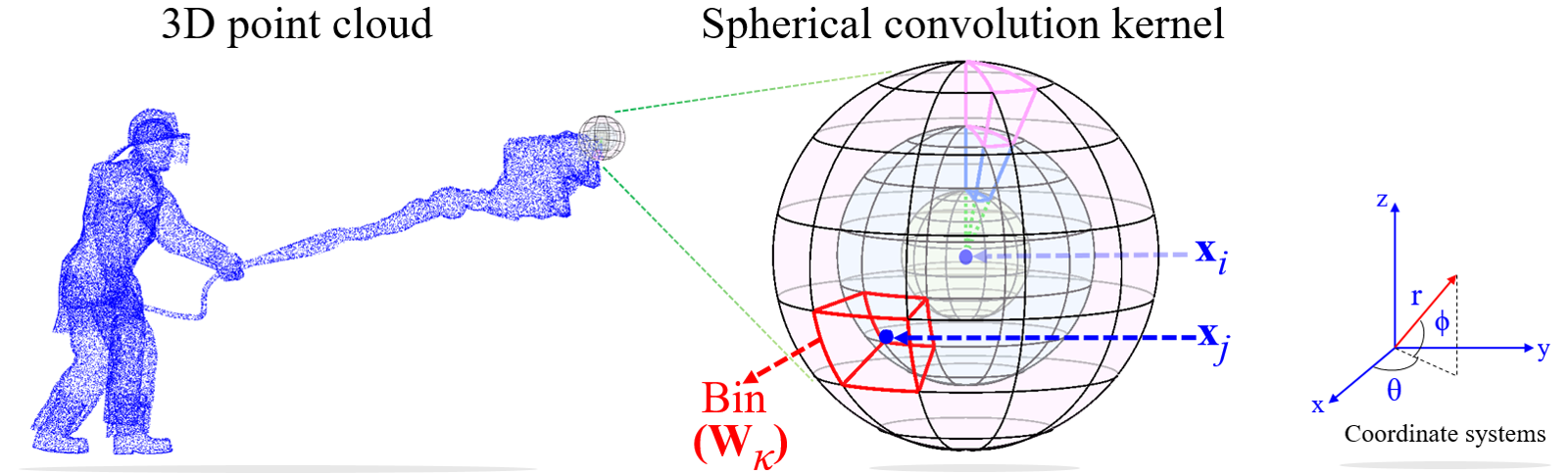}
  \caption{Illustration: The kernel processes raw point clouds by systematically splitting the space around a target point ${\bf x}_i$ into multiple \emph{bins}. For the $j^{\text{th}}$ neighboring point ${\bf x}_j$, it first determines its relevant bin and uses the weight matrix ${\bf W}_{\kappa}$ defined for that bin to compute activation value. }
  \label{fig:sphConv}
  \vspace{-3mm}
\end{figure}

Given an arbitrary point cloud $\mathcal{P}=\{\mathbf{x}_i\in \mathbb{R}^3\}_{i=1}^m$, we define the convolution kernel with the help of a sphere of radius $\rho\in \mathbb{R}^+$. For a target point $\mathbf{x}_i$, we consider its neighborhood $\mathcal{N}(\mathbf{x}_i)$ to comprise the points within the sphere centered at $\mathbf{x}_i$,  i.e. $\mathcal{N}(\mathbf{x}_i)=\{\mathbf{x}:d(\mathbf{x},\mathbf{x}_i)\leq \rho\}$, where $d(.,.)$ is a distance metric\footnote{We use the $\ell_2$ distance in this work.}.
We divide the sphere into $n \times p \times q$ \emph{`bins'} (see Fig.~\ref{fig:sphConv}) by partitioning the occupied space uniformly along the azimuth ($\theta$) and elevation ($\phi$) dimensions. We allow the partitions along the radial dimension to be non-uniform because cubic volume growth for large radius values can become undesirable. 
Our quantization of the spherical region is mainly inspired by 3DSC [3].
We also define an additional bin corresponding to the origin of the sphere to allow the case of self-convolution of points.
For each bin, we define a weight matrix $\mathbf{W}_{\kappa \in \{0, 1,\dots,n \times p \times q \}}\in\mathbb{R}^{s\times t}$ of learnable parameters, where $\mathbf{W}_0$ relates to self-convolution.  
Together, the $n \times p \times q + 1$ weight matrices specify a single spherical convolution kernel.


To compute the activation value of a target point $\mathbf{x}_i$, we first identify the relevant weight matrices of the kernel for each of its neighboring points $\mathbf{x}_j\in \mathcal{N}(\mathbf{x}_i)$. 
It is straightforward to associate $\mathbf{x}_i$ with $\mathbf{W}_0$ for self-convolution. For the non-trivial cases, we first represent the neighboring points in terms of their spherical coordinates that are referenced using $\mathbf{x}_i$ as the origin. That is, for each $\mathbf{x}_j$ we compute $\mathcal T(\boldsymbol{\Delta}_{ji})\rightarrow \boldsymbol{\psi}_{ji}$, where $\mathcal T(.)$ defines the transformation from Cartesian to Spherical coordinates and  $\boldsymbol{\Delta}_{ji}=\mathbf{x}_j-\mathbf{x}_i$.
Assuming that the bins of the quantized sphere are indexed by $k_\theta$, $k_\phi$ and $k_r$ along the azimuth, elevation and radial dimensions respectively, the weight matrices associated with the spherical kernel bins can be indexed as $\kappa = k_\theta + (k_\phi-1)\times n + (k_r-1)\times n\times p$, where $k_\theta \in \{1,\dots,n\},~k_\phi \in \{1,\dots,p\},~k_r \in \{1,\dots,q\}$. Using this indexing, we assign  each $\boldsymbol{\psi}_{ji}$; and hence $\mathbf{x}_j$ to its relevant weight matrix. 
In the $l^{\text{th}}$ network layer, the activation for the $i^{\text{th}}$ point is eventually computed as:
\begin{align}\label{spatio-conv}
  &\bold{z}^l_i  = \frac{1}{|\mathcal{N}(\bold{x}_i)|}\sum\limits_{j = 1}^{|\mathcal{N}(\bold{x}_i)|}\bold{W}^l_{\kappa}\bold{a}^{l-1}_j+\bold{b}^l,\\
  &\bold{a}^l_i = f(\bold{z}^l_i),
  \end{align}
where $\bold{a}^{l-1}_j$ is the activation value of a neighboring point from layer $l-1$, $\bold{W}^l_{\kappa}$ is the weight matrix, $\bold{b}^l$ is the bias vector, and $f(\cdot)$ is the non-linear activation function - ReLU [17] in our experiments.

To further elaborate on the characteristics of the spherical convolution kernel, let us denote the \emph{edges} of the kernel bins along $\theta$, $\phi$ and $r$ dimensions as the following: 
\begin{align}
\notag
&\boldsymbol{\Theta}=[\Theta_1,\dots,\Theta_{n+1}], ~\Theta_k<\Theta_{k+1}, {\Theta}_k\in[-\pi,\pi], \\
\notag
&\boldsymbol{\Phi}=[\Phi_1,\dots, \Phi_{p+1}]\big],~\Phi_k<\Phi_{k+1}, {\Phi}_k\in\big[-\frac{\pi}{2}, \frac{\pi}{2}] \\
\notag
&\bold{R}=[R_1,\dots,R_{q+1}], ~~R_k<R_{k+1}, R_k\in(0,\rho].
\end{align}
Under the constraint of uniform splitting along the azimuth and elevation, we have ${\Theta}_{k+1}-{\Theta}_k=\frac{2\pi}{n}$ and ${\Phi}_{k+1}-{\Phi}_k=\frac{\pi}{p}$.


\noindent {\bf Lemma 2.1:} \emph{If~$\Theta_k\cdot\Theta_{k+1}\geq 0$,  $\Phi_k\cdot\Phi_{k+1}\geq 0$ and $n>2$, then for any two points $\bold{x}_a\neq\bold{x}_b$ within the spherical convolution kernel, the weight matrices $\mathbf W_{\kappa}, \forall \kappa>0$, are applied asymmetrically.}\\
\noindent {\it Proof:} 
Let $\boldsymbol{\Delta}_{ab}= {\bf x}_a - {\bf x}_b =[\delta_x,\delta_y,\delta_z]^{\intercal}$, then $\boldsymbol{\Delta}_{ba}=[-\delta_x,-\delta_y,-\delta_z]^{\intercal}$.  Under the Cartesian to Spherical coordinate transformation,  we have $\mathcal T(\boldsymbol{\Delta}_{ab}) =\boldsymbol{\psi}_{ab}=[\theta_{ab},\phi_{ab},r]^{\intercal}$,  and $\mathcal T(\boldsymbol{\Delta}_{ba}) = \boldsymbol{\psi}_{ba}=[\theta_{ba},\phi_{ba},r]^{\intercal}$. Assume that the resulting $\boldsymbol{\psi}_{ab}$ and $\boldsymbol{\psi}_{ba}$ fall in the same bin indexed by $\kappa \leftarrow (k_\theta,k_\phi,k_r)$, i.e. ${\bf W}_{\kappa}$ will have to be applied symmetrically to the original points. 
In that case, under the inverse transformation $\mathcal T^{-1} (.)$, we have $\delta_z=r\sin\phi_{ab}$ and $(-\delta_z)=r\sin\phi_{ba}$.
The condition $\Phi_{k_\phi}\cdot\Phi_{k_\phi+1}\geq 0$ entails that $-\delta_z^2 = \delta_z\cdot(-\delta_z)=(r\sin\phi_{ab})\cdot(r\sin\phi_{ba})=r^2(\sin\phi_{ab}\sin\phi_{ba})\geq 0\Longrightarrow\delta_z=0$. Similarly, 
$\Theta_{k_\theta}\cdot\Theta_{k_\theta+1}\geq 0 \Longrightarrow \delta_y=0$.
Since $\bold{x}_a\neq\bold{x}_b$, for  $\delta_x\neq0$ we have $ \cos\theta_{ab} = -\cos\theta_{ba} \Longrightarrow |\theta_{ab}-\theta_{ba}|=\pi$. 
However, if  $\theta_{ab}$, $\theta_{ba}$ fall into the same bin, we have $|\theta_{ab}-\theta_{ba}|=\frac{2\pi}{n}<\pi$, which entails $\delta_x = 0$.  Thus, ${\bf W}_{\kappa}$ can not be applied to any two points symmetrically  unless both points are the same. 



The asymmetric characteristics of the spherical convolution kernel is significant because it restricts the sharing of the same weights between point pairs, which facilitates in learning more effective  features with finer geometric details. 
Lemma~2.1 also provides guidelines for the division of the convolution kernel into bins such that the asymmetry is always ensured.  It is worth mentioning that the weight matrices in the standard CNN kernels are also applied to pixel pairs asymmetrically. 


\paragraph{Relation to 3D-CNN}
Pioneering works in exploiting CNNs for 3D data rasterize 
the raw data into uniform voxel grids, and then
extract features using 3D-CNNs from the resulting volumetric representations [8], [15]. In 3D-CNNs, the convolution kernel of size $3\times3\times3=27$ is prevalently used, that  splits the space in 1 cell/voxel for radius $r=0$ (self-convolution); 6 cells for radius $r=1$; 
  12 cells for radius $r=\sqrt{2}$; and 8 cells for radius $r=\sqrt{3}$.
An analogous spherical convolution kernel for the same region can be specified with a radius $\rho=\sqrt{3}$, using the following edges for the bins: 
 \begin{align}\label{split-sphConv}
 \boldsymbol{\Theta}=[-\pi,-\frac{\pi}{2},0,\frac{\pi}{2},\pi];~
\boldsymbol{\Phi} =[-\frac{\pi}{2},-\frac{\pi}{4},0,\frac{\pi}{4},\frac{\pi}{2}]; ~\bold{R} = [\epsilon,1,\sqrt{2},\rho], \epsilon\rightarrow0^+. 
  \end{align} 
  This division results in a \emph{kernel size} (i.e. total number of bins) $4\times 4\times 3 + 1=49$, which is the coarsest multi-scale quantization allowed by Lemma~2.1. 
  
  Notice that, if we move radially from the center to periphery of the spherical kernel, we encounter identical number of bins (16 in this case) after each edge defined by $\bold{R}$, where fine-grained bins are located close to the origin that can encode detailed local geometric information of the points. This is in sharp contrast to 3D-kernels that  must keep the size of all cells constant and rely on increased input resolution of data to capture the finer details - generally entailing memory issues.  
The multi-scale granularity of spherical kernel makes them a natural choice for raw point clouds. Similar to 3D-CNN kernels, the spherical kernel also preserves the properties of translation-invariance and asymmetry.

 \begin{figure}
  \centering
 {\includegraphics[height=35mm]{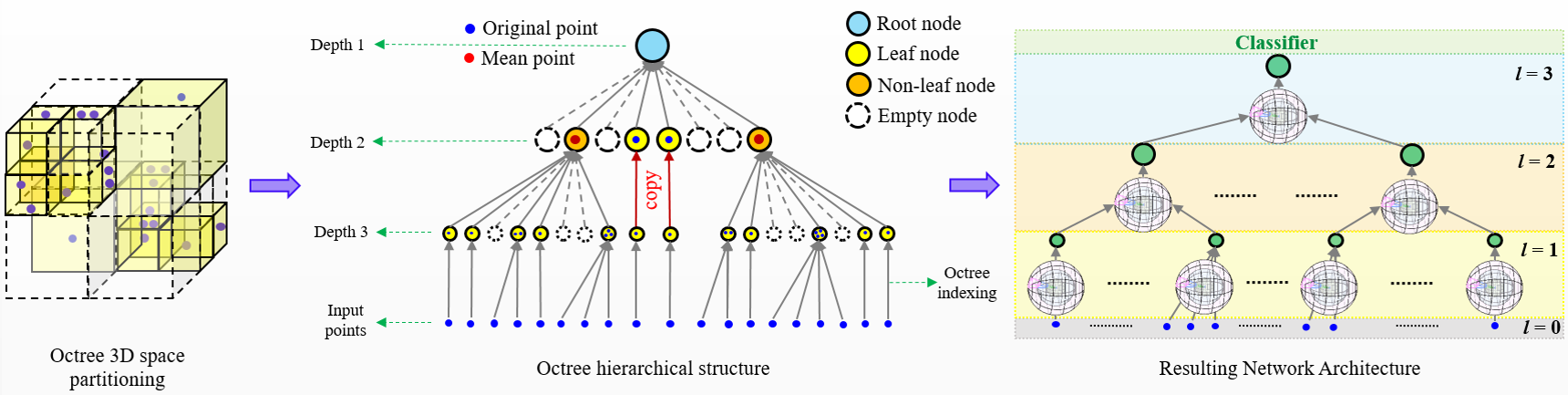}\label{octree}}
  \caption{Illustration of octree guided network architecture using a toy example: The point cloud in 3D space  is partitioned under an octree of depth 3. The corresponding tree representation allocates points to nodes at the maximum depth based on the  space partitioning, and computes the location of each parent node as the Expected location of its children. Leaf nodes on shallow branches are replicated to match the maximum depth.  The corresponding neural network has the same number of hidden layers as tree depth, and it learns spherical convolution kernels for feature extraction.}
  \label{fig:network}
  \vspace{-3mm}
\end{figure}
 \vspace{-2mm}
\section{Network Architecture}
\vspace{-2mm}
Recent years have seen few attempts to directly process point clouds with neural networks,  e.g. PointNet++ [10], ECC [12]. These works predominately rely on K-NN  or range searches to build local regions around the points to perform operations like convolution. However, to process large point clouds these search strategies become prohibitively expensive.   In the 3D community, a popular method to efficiently partition the data points into their approximate local neighborhoods is tree-structuring, e.g. Kd-tree [22].  The hierarchical nature of tree structures also provide guidelines for the neural network architectures that can be used to process the structured point clouds. More importantly, the structured data also possess
the much desired attributes of permutation and translation invariance for direct neural network processing.

In this work, we use the octree structure [21] for representing point clouds and design the neural network architecture based on the resulting trees. Our choice of using octrees is inspired by their natural amenability to neural networks as the base data structure~[11],  and their ability to account for more data in point neighborhoods as compared to e.g.~Kd-trees.  An illustration of partitioning the 3D space under octree structure, the resulting tree, and the construction of neural network using our strategy is shown in Fig.~\ref{fig:network} for a toy example. For an input point cloud $\mathcal{P}$, we  construct an octree of depth $L$  ($L = 3$ in the figure). In this construction, the splitting of leaf nodes is fixed to use a maximum node capacity of one point, which ensures that all leaf nodes contain only one point, with the exception of the last layer nodes. The allocation of multiple points in the last layer leaf nodes directly results from the allowed finest partitioning of the space. For the sub-volumes in 3D space that are not densely populated with points, our splitting strategy can result in leaf nodes before the tree reaches its maximum depth. In such cases, to facilitate mapping of the tree structure to a neural network, we simply replicate such leaf nodes to the maximum depth of the tree. 
We safely ignore the empty leaf nodes while implementing the network architecture resulting in computational and memory benefits.

Based on the hierarchical tree structure, our network architecture has $L$ hidden layers. Notice that, in Fig.~\ref{fig:network},  we use $l = 1$ for the first hidden layer that  corresponds to Depth $= 3$ for the tree. We will use the same convention in the text to follow. For each non-empty node in the tree, there is a corresponding neuron in our neural network.  Recall that, a spherical convolution kernel is specified with a target point over whose neighborhood the convolution is performed. Therefore, to facilitate the  spherical convolutions, we associate a 3D location with each neuron. These locations are computed as the Expected values of the children of each node in the octree, except for the leaf nodes at the maximum tree depth. For those nodes, the associated locations are simply the mean values of the data points.  
A neuron uses its associated location to select the appropriate spherical kernel and later applies the non-linear activation (not shown in Fig.~\ref{fig:network} for brevity).  In our network, all the (spherical) convolutions before the last layer are followed by batch normalization and ReLU activations. The  convolution operation in the last layer is followed by ReLU only.

Let us denote the location associated with the $i^{\text{th}}$ neuron in the $l^{\text{th}}$ layer of the network as $\bold{\bar{x}}_i^l$. From $l =1$ to $l = L$, we can  represent the locations associated with all neurons as $\mathcal{Q}^1=\{\bold{\bar{x}}_i^1\}_{i=1}^{m_1}$, $\dots$, $\mathcal{Q}^L=\{\bold{\bar{x}}_1^L\}_{i = 1}^{m_L}$. Denoting the raw input points as $\mathcal{Q}^0=\{\bold{\bar{x}}_i^0\}_{i=1}^{m_0}$, $\bold{\bar{x}}_i^l$ is numerically computed as:
\begin{align}\label{node-xyz}
  \bold{\bar{x}}^l_i = \frac{\sum\limits_{\bold{\bar{x}}_j^{l-1}\in \mathcal{N}(\bold{\bar{x}}^l_i)}\bold{\bar{x}}_j^{l-1}}{|\mathcal{N}(\bold{\bar{x}}^l_i)|},
  \end{align}
  where $\mathcal{N}(\bold{\bar{x}}_i^l)$ contains the locations of relevant children nodes in the octree.  It is worth noticing that the strategy used  for specifying the network layers also entails that $|\mathcal Q^{l-1}| > |\mathcal Q^l|$,  to the end that there is only one location associated with the final hidden layer corresponding to the root of the octree. 
Thus, from the first to the last layer, the features learned by the network move from lower to higher level of abstraction similar to the standard CNN architectures.

Associating the spherical nature of the neighborhood relation considered in our network to the cubic partitioning of the space by octree, a subtle detail is worth mentioning. 
Consider that $\bold{x}_{\min}$ = $[x_{\min}, y_{\min}, z_{\min}]^{\intercal}$, and $\bold{x}_{\max}$ = $[x_{\max}, y_{\max}, z_{\max}]^{\intercal}$  define the range of coordinates of points in a given cubic cell of space partitioning. Depending on the network layer, the spherical neighborhood associated  with a corresponding neuron is of radius $\rho=2^{l-L-1}\cdot d(\bold{x}_{\min},\bold{x}_{\max})$, which may not strictly circumscribe all point of the cubic volume. In practice, the number of such 'outliers' is minuscule. Nevertheless, we also take those points into account by assigning them to the outer-most bins of convolution kernels based on their azimuth and elevation values. 

Our neural network performs inter-layer convolutions instead of intra-layer convolutions. This drastically reduces the operations required to process large point clouds when compared with graph-based networks [1], [4], [12], [16], [20]. We note that for all nodes with a single child, only self-convolutions are performed in the network. It is also worth mentioning that due to its unconventional nature, spherical convolution operation is not readily implemented using the existing deep learning libraries, e.g. matconvnet~[14]. Therefore, we implement it ourselves with CUDA C++ and mex interface. For the other modules such as ReLU, batch normalization etc., we use matconvnet. 


\vspace{-2mm}
\section{Comparison with Related Work}
\vspace{-2mm}

PointNet [9] first processes point clouds directly using  $x,y,z$ coordinates as input features. The network learns point-wise features with shared MLPs, and extracts a global feature with max pooling. The main limitation of PointNet is that it explores no geometric context in the point-wise feature learning. This was later addressed by the authors of PointNet++ [10] by applying max-pooling to the local regions hierarchically.  This enhancement builds local regions  using K-NN search as well as range searches. However, both PointNets [9], [10] always aggregate context information with max pooling, and no convolution modules are explored in the networks. 
In regards to processing point clouds with deep learning framework using tree structures, Kd-network [5]  is the first prominent contribution. Kd-network also uses point coordinates as the input, and computes  the feature of a parent node by concatenating the features of children in a \emph{balanced} tree. However, the performance of Kd-network depends on elaborate randomness of the splitting dimensions in the tree construction. In contrast, our approach relies on deterministic geometric relationships between points. The major similarity between our approach, Kd-network and PointNets comes in terms of directly accepting the spatial coordinates of points as the input features. From the perspective of convolutional networks for 3D data, the following two categories of networks also  relate to our approach.

\vspace{-1mm}
\subsection{Graph Convolutional Networks}
\vspace{-1mm}
Graph convolutional networks can be grouped into spectral networks [1], [4], [20] and spatial networks [12]. The spectral networks perform convolution in the spectral domain relying on the graph Laplacian and adjacency matrices, while the spatial networks perform convolution in the spatial domain. 
A major limitation of spectral networks is that they demand the graph structure to be fixed, which makes their applications to the data with varying graph structures (e.g.~point clouds) challenging. Yi \emph{et al.} [16] attempted to address this issue with Spectral Transformer Network (SpecTN), similar to STN [18] in the spatial domain. However, the signal transformation from spatial to spectral domains and vice versa requires computations of complexity $\mathcal{O}(n^2)$. 
ECC [12] is the first work for point cloud analysis with graph convolution in the spatial domain. The authors adopted MLPs to generate convolution filters between the connected vertices dynamically. 
However, dynamic generation of the edge filters introduces both convolution delay and extra computation burden. In addition, the graph construction and coarsening is dependent on range searches, which remains  inefficient. Our approach builds the network architecture directly from the octree avoiding altogether the requirement for graph construction and coarsening. Moreover, the spherical convolution  effectively explores the geometric context of each point in our method without the need of dynamic filter generation.

\vspace{-1mm}
\subsection{3D Convolutional Neural Networks} 
\vspace{-1mm}
3D-CNN kernels are  applied to volumetric representations of 3D data. In the earlier works, only low input resolution point clouds could be processed, e.g. 30$\times$30$\times$30 [15], 32$\times$32$\times$32 [8]. Because for dense data, the computational and storage costs of the network grows cubically with the input resolution. 
Different solutions have been proposed to address these issues e.g. [2], [7]. Most recently, 
Riegler \emph{et al.} [11] proposed OctNet, that represents point clouds with a hybrid of shallow grid octrees (depth=3). Compared to its dense peers, OctNet reduces the computational and memory costs to a large degree, and is applicable to high-resolution inputs up to 256$\times$256$\times$256. 
Although our architecture is also based on octrees, it significantly differs from OctNet. Firstly,  in OctNet, the nodes are treated as mere voxels and 3D-CNN kernel is used to process the resulting representation, which is in sharp contrast to our spherical treatment of the features. Secondly, our network is based on a single deep octree rather than a hybrid of shallow octrees, as in OctNet.   Another difference between the two networks is that our network accepts $x,y,z$ coordinates of points as inputs, whereas OctNet requires binary occupancy features. Furthermore, OctNet processes all octree nodes whereas our network is based on a trimmed version of the tree that discounts the empty nodes for efficiency.

\vspace{-2mm}
\section{Experiments}
\vspace{-2mm}
We focus on the task of 3D object classification to evaluate the proposed spherical convolutional neural network. The evaluation is conducted on ModelNet10 and ModelNet40 benchmark datasets~[15]. These standard data sets are created from clean CAD models.  The ModelNet10 contains 10 categories of aligned object meshes, and  the samples are split into 3,991 training examples and 908 test samples. ModelNet40 contains object meshes for 40 categories with 9,843/2,468 training/testing samples split. We use the non-aligned version of ModelNet40  for the sake of variety.


Our network can consume large input 3D models, hence we train it with the input size of 30K points. Due to octree structuring it also has the ability of getting trained and tested on data samples of different sizes. Therefore, we also examine the networks trained on 30K point samples and tested on larger samples, i.e.  40K and 50K points. We keep the depth of the octree fixed to 8 levels in our experiments for a uniform evaluation. The number of feature channels used are 32-32-64-64-128-128-256-512. A classifier with a  fully connected layer followed by softmax is trained in an end-to-end fashion with our network. We perform training with  Stochastic Gradient Descent with starting learning rate 0.1 and 0.0005 momentum. The learning rate decays by a factor of 10 after first 50 epochs, and keeps decaying with the same factor after every 20 epochs thereon, until all 100 epochs are complete. Training is conducted on Titan Xp GPU with 12 GB memory. We use batch size 16 in all experiments. 
The hyper-parameters of the network were empirically optimized using cross-validation. 

There are instances in literature where normals are also exploited as input features for accuracy advantage. However, normals may not always be readily available for point clouds, and their use also accompanies undesired computational overhead in the forward pass. Therefore, we restrict the input features of our network to contain only the $(x,y,z)$ coordinate values of  points.  To standardize the input, we normalize the 3D point coordinates by fixing the origin of the point cloud at its center of mass and rescaling all points to fit within a cube of $[-1,1]^3$.
For the convolution kernels,  we use the size of $8\times2\times3+1$, in which the radial dimension is split uniformly. 

We abbreviate our approach as $\Psi$-CNN\footnote{A Greek alphabet is chosen as prefix to avoid duplication with other SCNNs, e.g. [23], [24].} in 
Table~\ref{compare2Others} that summarizes the object recognition performance on the ModelNets. 
Along the `class' and `instance' accuracies, where the former is the average accuracy per object category, we also report the input feature type and the number of network layers - counting only the layers for global feature representation. 
As can be seen, the $\Psi$-CNN achieves accuracies competitive with a variety of networks for point cloud processing.
For ModelNet10, Kd-network performs slightly better than our network when it uses twice the number of layers used by our network, which accompanies with a training process of 5 days [5]. In terms of network architecture, OctNet is the most relevant. The $\Psi$-CNN achieves significantly better performance than OctNet with almost half the number of layers. We can attribute this gain mainly to the accurate exploration of geometric information by spherical kernels for which we tailor our network. Recall that, among the compared networks, only ECC and OctNet are able to perform convolutions. The proposed convolutional network is able to significantly outperform ECC despite the later having dense intra-layer connections. In its current state, the PointNets seem to outperform our approach. Nevertheless, the ability of successfully applying convolutions to unstructured point clouds promises much better prospects for our method.  

Similar to other methods, we also take advantage of data augmentation in the results reported in Table~\ref{compare2Others}. We perform random sub-sampling, random azimuth rotation (up to $\frac{\pi}{6}$ rad) and noisy translation (std.~dev = 0.02) to increase the number of train and test examples five times of the original. In Table~\ref{tab:augment}, we summarize the results of $\Psi$-CNN with and without augmentation for testing the network trained on 30K size point clouds but using test samples of different sizes. For 50K, we do not perform random sub-sampling augmentation as we set 50K to be the maximum number of points.  


\vspace{-1mm}
\subsection{Scalability}
\begin{figure}[!t]
\includegraphics[width=42mm]{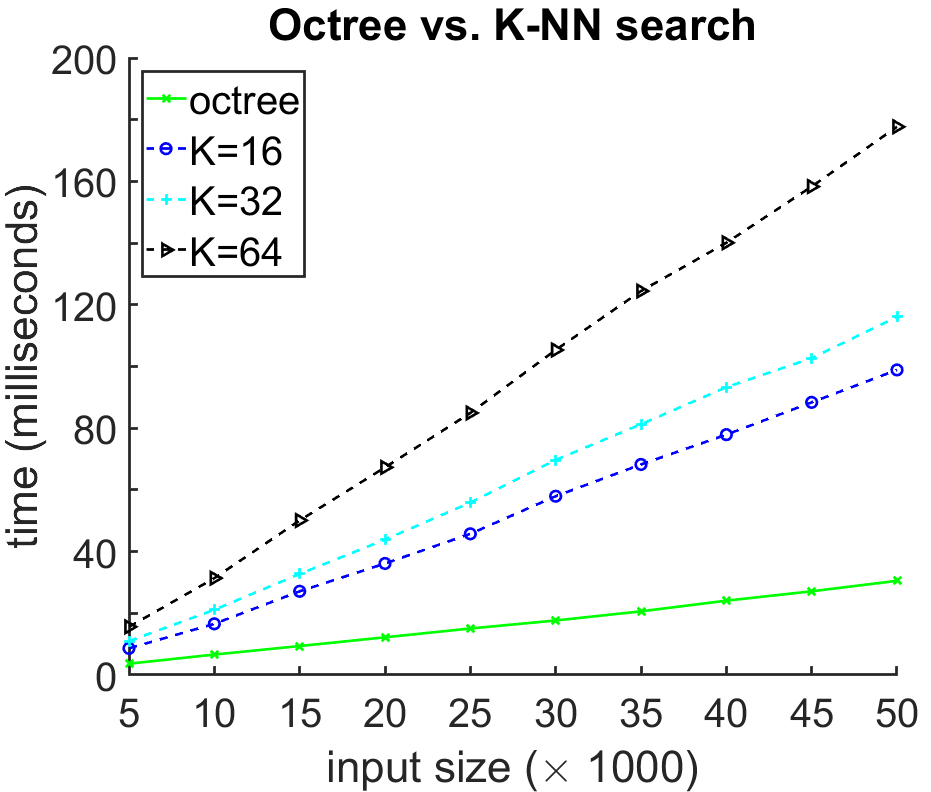}
\hspace{1mm}
\includegraphics[width=40mm]{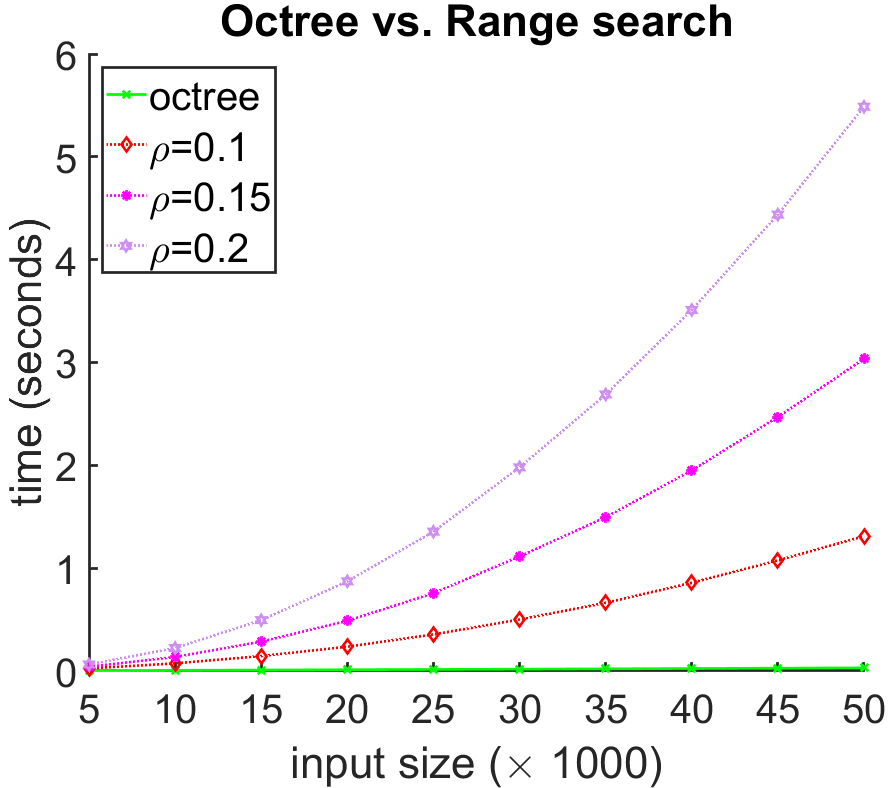}
\hspace{1mm}
\includegraphics[width=42mm]{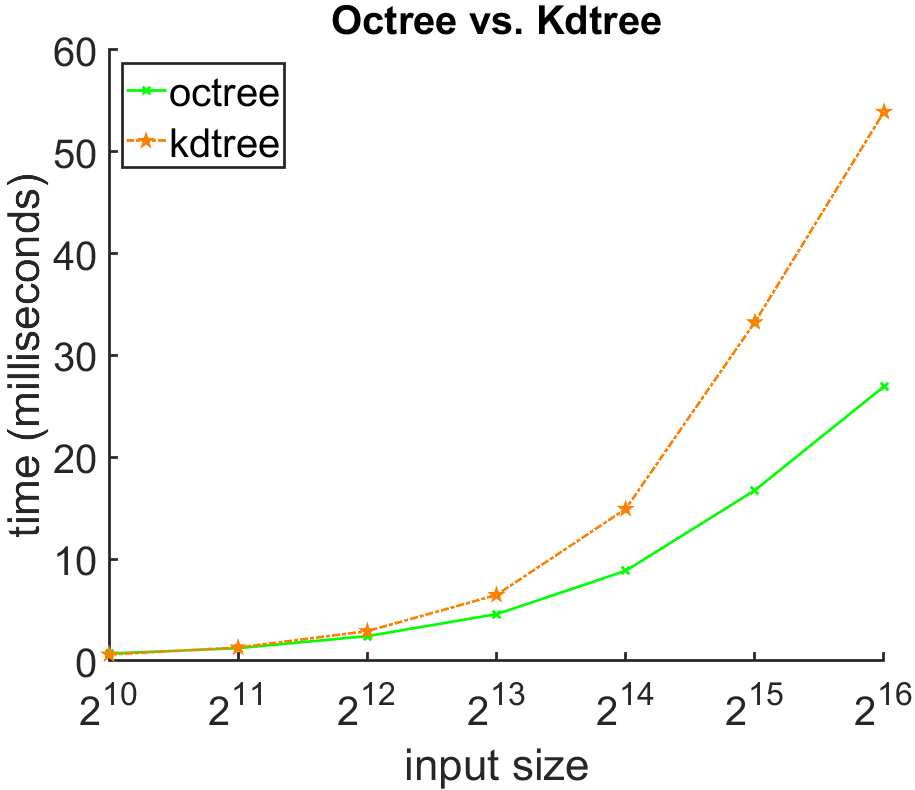}
\caption{Comparison of octree structuring with K-NN, range search and Kd-tree for neighborhood computation. Better scalability of octree leads to better scalability of our approach. }\label{time-analysis}
\label{fig:graphs}
\end{figure}

\vspace{-1mm}
For geometrically meaningful convolutions, knowledge of local neighborhood of points is imperative. ECC [12] exploits range search for this purpose. Another obvious choice is K-NN clustering. For tree structures, e.g. Kd-tree, octree, this information is readily available under the partitioning of the space. In our approach, local neighborhoods must be computed for both training and testing samples which makes this computation a major factor in deciding the scalability of the approach. In Fig.~\ref{fig:graphs}, we compare the timings of computing  neighborhoods under different choices with octree construction. As can be seen, for larger number of input points, octree structuring scales very well as compared to K-NN and range searching. Moreover, its scaling is also better than Kd-tree for large input sizes because the binary partitioning in Kd-tree causes the tree depth to increase much more than the partitioning in octree. Thus, we can argue that the choice of octree structure makes our approach much more scalable than approaches using K-NN, range search or even Kd-tree.



In Table \ref{test_time}, we also report the test time of our network for point clouds of different sizes. For a given sample, the total test time consists of time required to construct octree and then performing the forward pass. 
The table also includes the time required to compute normals as a reference. Note that, our approach does not require to compute normals and the reported results are without taking benefit from normal information. The last column of Table \ref{test_time} justifies the computational reasons behind not exploiting normal information for better scalability.

\begin{table}
\centering
\begin{tabular}{l|c|c|c|c|c|c}
  \hline
  \multirow{ 2}{*}{Method}& \multirow{ 2}{*}{Input feature} & \multirow{ 2}{*}{Layers} & \multicolumn{2}{c|}{ModelNet10} & \multicolumn{2}{c}{ModelNet40} \\
  \cline{4-7}
 & & &class & instance & class & instance \\
  \hline
  \hline
  PointNet &$xyz$ & 2T-Nets+5 & -- & -- & 86.2 &89.2 \\
  PointNet++&$xyz$& 9& -- & -- & --  &90.7 \\
  PointNet++&$xyz$+normal&9 & -- & -- & --  &91.9 \\
  Kd-network &$xyz$&11 & 92.8 &93.3 &86.3 &90.6\\
  Kd-network &$xyz$&16 & 93.5&94.0& 88.5&91.8\\
  OctNet &0/1&15 &90.1 & 90.9 & 83.8 & 86.5 \\
  ECC & 0& 5 &90.0&90.8 & 83.2&87.4 \\
  \hline
   $\Psi$-CNN (proposed) & $xyz$ & 8 & 93.2 & 93.6 & 85.2  & 89.7  \\
  \hline
\end{tabular}
\caption{Classification performance on ModelNets [15].} \label{compare2Others}
\vspace{-3mm}
\end{table}

\begin{table}
\centering
\begin{tabular}{l|c|c|c|c}
  \hline
  \multirow{ 2}{*}{Configuration}& \multicolumn{2}{c|}{ModelNet10} & \multicolumn{2}{c}{ModelNet40} \\
  \cline{2-5}
 & class & instance & class & instance \\
  \hline
  \hline
  $\Psi$-CNN+30K+no-augment & 90.3
  &91.2
  & 82.6&87.2 \\
  $\Psi$-CNN+40K+no-augment &90.7 & 91.5 & 82.9& 87.4\\
  $\Psi$-CNN+50K+no-augment &91.0 & 91.6 & 82.8& 87.4\\
  \hline
  $\Psi$-CNN+30K+augment &92.8&93.3  & 85.2 & 89.7\\
  $\Psi$-CNN+40K+augment &93.1 & 93.5 & 85.1& 89.7\\
  $\Psi$-CNN+50K+augment &93.2 & 93.6 & 84.9& 89.6\\
  \hline
\end{tabular}
\caption{Recognition performance with and without data augmentation. The training sample size is kept fixed to 30K points, and test sample sizes (shown) are varied.} \label{tab:augment}
\vspace{-3mm}
\end{table}
\begin{table}
\centering
\begin{tabular}{c|c|c|c||c}
  \hline  
 Point Cloud size & Octree construction & Forward pass&~~~~~Total~~~~~& \cellcolor{red!15} Normal computation \\
  \hline
  30K & 18 & 110 & 128& \cellcolor{red!7}100\\
  \hline
  40K &  26 & 140 & 166& \cellcolor{red!7}136\\
  \hline
  50K & 31 & 172 & 203& \cellcolor{red!7}173\\
  \hline
\end{tabular}
\caption{Per sample test time (ms) for inputs of different sizes. The per-sample time for computing normals is shown for reference only - indicated by red. Our approach does not compute normals. }
\label{test_time}
\end{table}

\subsection{Visualization}
\begin{figure}[!t]
\centering
\hspace{-5mm}
 \subfigure[Point cloud evolution]{
 \begin{tabular}{cccccc}
     $l=1$ & $l=2$ & $l=3$ & $l=4$ & $l=5$ & $l=6$ \\
   \includegraphics[height=12mm]{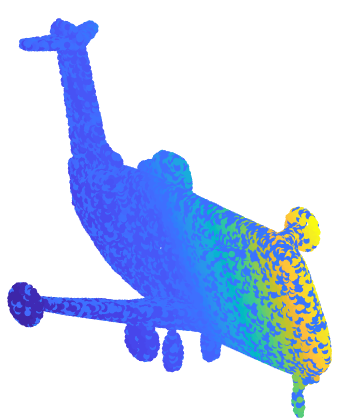}   & \includegraphics[height=12mm]{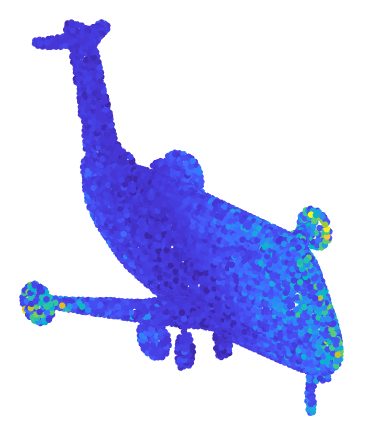} & \includegraphics[height=12mm]{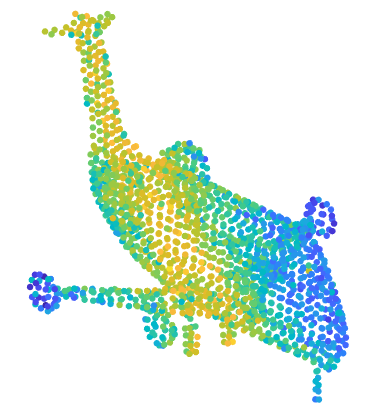} & \includegraphics[height=12mm]{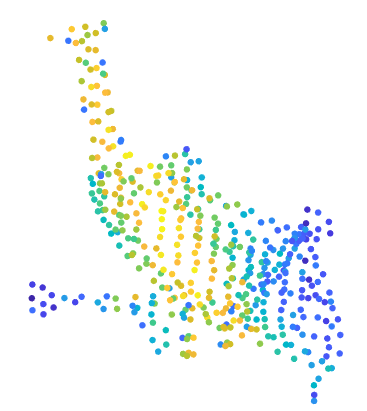} & \includegraphics[height=12mm]{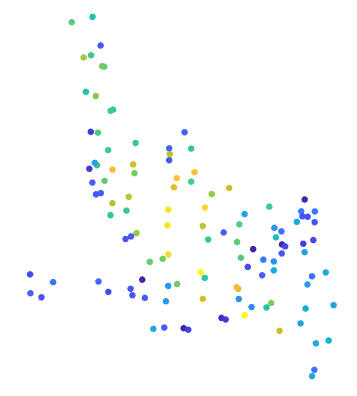} & \includegraphics[height=12mm]{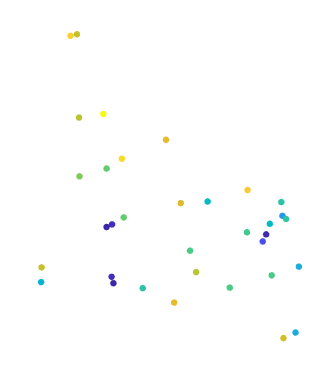} \\
     \includegraphics[height=12mm]{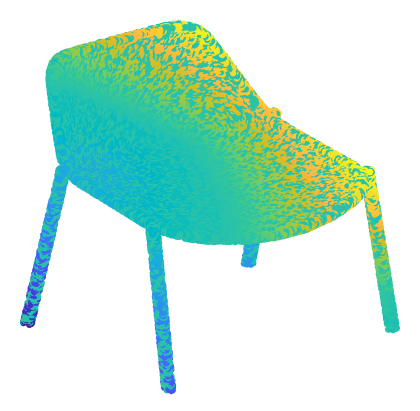} & \includegraphics[height=12mm]{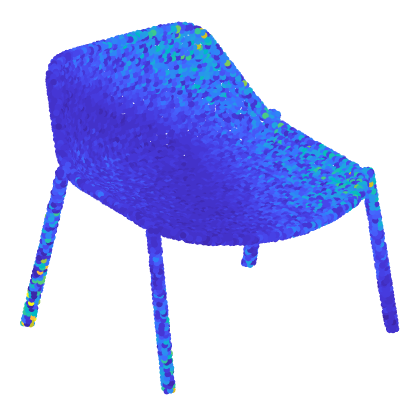} & \includegraphics[height=12mm]{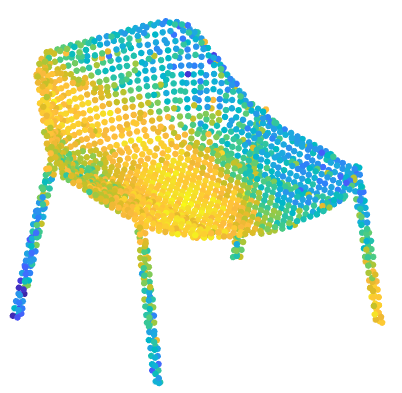} & \includegraphics[height=12mm]{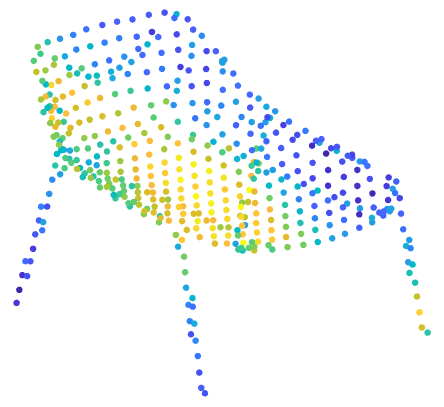} & \includegraphics[height=12mm]{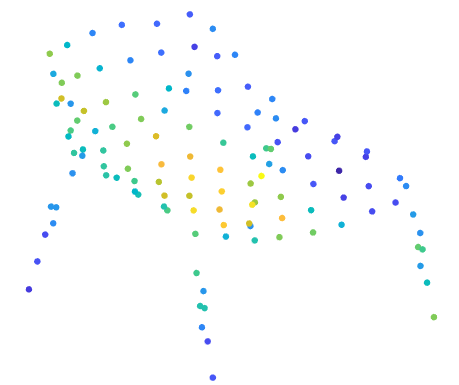} & \includegraphics[height=12mm]{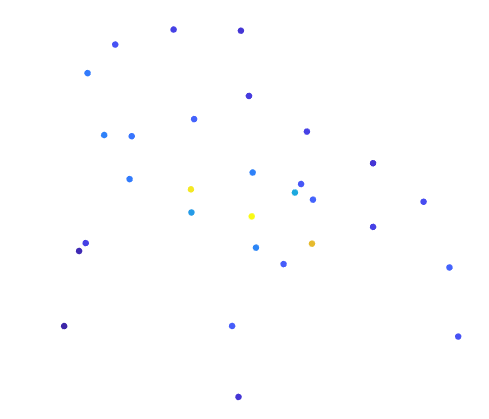} \\
   \end{tabular}}
   \hspace{4mm}
 \subfigure[Spherical kernel]{\begin{tabular}{c}
 \includegraphics[height=17.5mm]{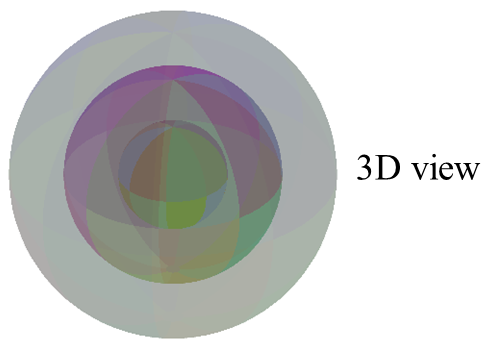}\\
 \includegraphics[height=17mm]{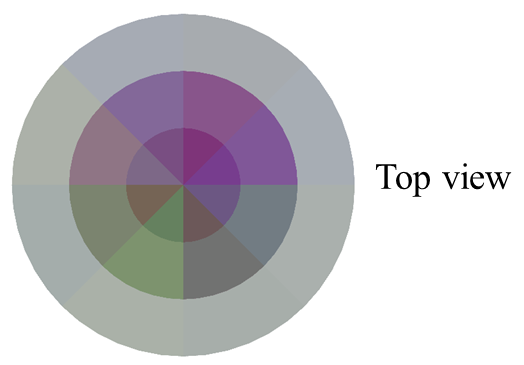}\\
\end{tabular}}
\caption{(a) Evolution of the point cloud and its representation on the tree structure.
Similar colors corresponds to similar feature representations. (b) One pattern learned by the spherical kernel is shown with a 3D view and a top view.}\label{visualize}
\end{figure}

We show the evolution of point clouds and their feature representation with two representative examples in Fig.~\ref{visualize}. 
It can be seen that as the layer number increases, the point cloud becomes coarser and its feature representation becomes more distinctive. In the higher layers, the point cloud loosely becomes a skeleton representation of the original point cloud. We also visualize a representative example of spherical kernel learned on ModelNet10 with a 3D view and the top view. 
For further visualizations of convolution kernels and point cloud evolutions we refer to the supplementary material submitted with the paper.


\section{Conclusion }
We introduced the concept of spherical convolutional kernel and demonstrated its utility with a neural network architecture guided by octree partitioning of the raw point clouds. The network successively applies the convolution kernels in the neighborhood of  its neurons. These locations are governed by the nodes of the underlying octree. To perform the convolutions, the spherical kernel partitions its occupied space into multiple bins and associates a  weight matrix to each bin. These matrices are learned with network training. It is shown that the resulting spherical convolutional neural network can effectively  process irregular 3D point clouds in a scalable manner, achieving very promising results for the task of model recognition.
Other hierarchical structures compatible with  spherical convolution will be explored in the future.

\subsubsection*{Acknowledgments}
This research was supported by ARC grant DP160101458. The Titan Xp used for this research was donated by NVIDIA Corporation.

\section*{References}
\medskip
\small
[1] M. Defferrard, X. Bresson, and P. Vandergheynst. Convolutional neural networks on graphs with fast localized spectral filtering. In {\it Advances in Neural Information Processing Systems}, pages 3844--3852, 2016.

[2] M. Engelcke, D. Rao, D. Zeng Wang, C. Hay Tong, and I. Posner. Vote3Deep: Fast object detection in 3D point clouds using efficient convolutional neural
networks. In {\it International Conference on Robotics and Automation}, 2017.

[3] A. Frome, D. Huber, R. Kolluri, T. B{\"u}low, and J. Malik.
Recognizing objects in range data using regional point descriptors. In {\it European Conference on Computer Vision}, pages 224--237, 2004.


[4] T. N. Kipf and M. Welling. Semi-supervised classification with graph convolutional networks. In {\it International Conference on Learning Representations}, 2017.

[5] R. Klokov and V. Lempitsky. Escape from cells: Deep kd-networks for the recognition of 3D point cloud models. In {\it Proceedings of the IEEE International Conference on Computer Vision}, 2017.

[6] Y. Lecun, L. Bottou, Y. Bengio, and P. Haffner. Gradient-based learning applied to document recognition. In {\it Proceedings of the IEEE}, pages 2278--2324, 1998.


[7] Y. Li, S. Pirk, H. Su, C. R. Qi, and L. J. Guibas. FPNN: Field probing neural networks for 3D data. In {\it Advances in Neural Information Processing Systems}, pages 307--315, 2016.

[8] D. Maturana and S. Scherer. VoxNet: A 3D convolutional neural network for real-time object recognition. In {\it IEEE International Conference on Intelligent Robots and Systems},
pages 922--928, 2015.


[9] C. R. Qi, H. Su, K. Mo, and L. J. Guibas. PointNet: Deep learning on point sets for 3D classification and segmentation. In {\it Proceedings of the IEEE Conference on Computer Vision and Pattern Recognition}, 2017.

[10] C. R. Qi, L. Yi, H. Su, and L. J. Guibas. PointNet++: Deep hierarchical feature learning on
point sets in a metric space. In {\it Advances in Neural Information Processing Systems}, pages 5105--5114, 2017.

[11] G. Riegler, A. O. Ulusoy, and A. Geiger. OctNet: Learning deep 3D representations at high resolutions. In {\it Proceedings of the IEEE Conference on Computer Vision and Pattern Recognition}, pages 6620--6629, 2017.


[12] M. Simonovsky and N. Komodakis. Dynamic edge-conditioned filters in convolutional neural networks on graphs. In {\it Proceedings of the IEEE Conference on Computer Vision and Pattern Recognition}, pages 29--38, 2017.

[13] F. Tombari, S. Salti, and L. D. Stefano. Unique signatures of histograms for local surface description. In {\it European Conference on Computer Vision}, pages 356--369, 2010.

[14] A. Vedaldi and K. Lenc. MatConvNet -- Convolutional Neural Networks for MATLAB. In {\it Proceedings of the 25th annual ACM International Conference on Multimedia}, 2015.

[15] Z. Wu, S. Song, A. Khosla, F. Yu, L. Zhang, X. Tang and J. Xiao. 3D ShapeNets: A deep representation for volumetric shapes. In {\it Proceedings of the IEEE Conference on Computer Vision and Pattern Recognition}, pages 1912--1920, 2015.


[16] L. Yi, H. Su, X. Guo, and L. J. Guibas. SyncSpecCNN: Synchronized spectral CNN for 3D shape segmentation. In {\it Proceedings of the IEEE Conference on Computer Vision and Pattern Recognition}, pages 6584--6592, 2017.

[17] V. Nair and G. E. Hinton. Rectified linear units improve restricted boltzmann machines. In {\it Proceedings of the 27th international conference on machine learning}, pages 807--814, 2010.

[18] M. Jaderberg, K. Simonyan, A. Zisserman, and K. Kavukcuoglu. Spatial transformer networks. In {\it Advances in Neural Information Processing Systems}, pages 2017--2025, 2015.

[19] A. X. Chang et al. Shapenet: An information-rich 3d model repository. {\it arXiv preprint arXiv:1512.03012}, 2015.

[20] J. Bruna, W. Zaremba,
A. Szlam, and Y. LeCun. Spectral Networks and Deep Locally Connected
Networks on Graphs. 
In {\it International Conference on Learning Representations}, 2014.

[21] D. Meagher. Geometric modeling using octree encoding.
In {\it Computer Graphics and Image Processing}, 1982.

[22] J. L. Bentley. Multidimensional binary search trees used
for associative searching. In{\it Communications of the ACM},
pages 509--517, 1975.

[23] B. Liu, M. Wang, H. Foroosh, M. Tappen, M. Pensky. Sparse Convolutional Neural Networks. In  {\it IEEE Conference on Computer Vision and Pattern Recognition}, pages 806-814.

[24] A. Parashar, M. Rhu, A. Mukkara, A. Puglielli, R. Venkatesan, B. Khailany, and W. J. Dally,  SCNN: An accelerator for compressed-sparse convolutional neural networks. In {\it 44th Annual International Symposium on Computer Architecture}, pages 27-40, 2017. 
\end{document}